\documentclass[11pt]{article}
\usepackage[preprint]{acl}
\usepackage{times}
\usepackage{latexsym}
\usepackage{amsmath}
\usepackage{amssymb}
\usepackage{booktabs}
\usepackage{graphicx}
\usepackage{microtype}
\usepackage{url}
\usepackage{xcolor}
\usepackage{multirow}

\title{AMEND: Constraint-Guided Multi-Agent Text Detoxification\\with Isolated Critics}

\author{Tianyu Wang \\
  School of Computer Science and Technology \\
  Soochow University \\
  \texttt{tywangcs@foxmail.com}}

\begin{document}
\maketitle

\begin{abstract}
Text detoxification requires balancing toxicity removal, semantic preservation, and fluency. Existing methods typically optimize an unconstrained scalar objective---enabling safety scores to mask semantic loss---or rely on open-ended iterative debate prone to semantic drift and high computational cost. We propose \textbf{AMEND}, an inference-time framework that decouples detoxification into role-conditioned proposal generation, isolated multidimensional criticism, and constraint-first selection. AMEND evaluates diverse candidates with role-blind critics and enforces strict semantic and fluency feasibility before ranking by utility. To maximize efficiency, \textbf{AMEND Lite} restricts subsequent search to at most one targeted repair round only when residual toxic spans remain. On ParaDetox, one-round selection over diverse roles improves the joint evaluation metric by $+0.0508$ over the best single role. On a frozen confirmation test set, AMEND Lite achieves a joint score of $0.6597 \pm 0.0062$, outperforming unconstrained multi-round loops by $+0.0489$ ($p < 0.05$) while reducing inference requests by $28.4\%$.
\end{abstract}

\section{Introduction}

Text detoxification---transforming abusive or toxic text into civil discourse while retaining the intended message---is an essential capability for safe human-AI interaction and online moderation. However, the task is fundamentally governed by a three-way tension between toxicity reduction, semantic preservation, and natural fluency \citep{mir-etal-2019-evaluating,logacheva-etal-2022-paradetox}. Aggressive deletion or masking easily sanitizes toxic lexical tokens but frequently dismantles the speaker's core intent or introduces jarring disfluencies. Conversely, conservative edits risk leaving subtle or contextual toxicity intact.

Existing methods generally fall into two paradigms. First, supervised sequence-to-sequence models and reward-guided decoders typically optimize a collapsed scalar objective \citep{dale-etal-2021-text,hallinan-etal-2023-detoxifying}. In practice, scalar averaging enables a high safety score to compensate for an unacceptable collapse in semantic fidelity---a failure mode where the model evades toxicity simply by altering the topic. Furthermore, generating a single candidate conceals valuable strategic diversity between conservative localized substitutions and holistic pragmatic reframing. Second, emerging agentic and iterative self-refinement methods \citep{madaan-etal-2023-self-refine,chen-etal-2024-reconcile} attempt multi-round editing. However, open-ended debate often suffers from feedback drift, role sycophancy, and unbounded token consumption with rapidly diminishing returns \citep{smit-etal-2024-going-mad}.

In this paper, we propose \textbf{AMEND} (\textbf{A}gentic \textbf{M}ulti-critic d\textbf{E}toxificatio\textbf{N} with constrainte\textbf{D} selection), an inference-time framework that structures detoxification as a decoupled proposal--critique--selection pipeline. Rather than forcing a single model to simultaneously generate, evaluate, and balance competing goals, AMEND disaggregates the process into three specialized stages:
\begin{enumerate}
    \item \textbf{Role-Conditioned Proposals}: A shared base language model is prompted under three complementary personas: a \texttt{minimal\_editor} performing precise span replacement, a \texttt{tone\_rewriter} converting vitriol into assertive, civil phrasing, and a \texttt{semantic\_paraphraser} restructuring the utterance while rigidly preserving facts and stance.
    \item \textbf{Isolated Multidimensional Critics}: Candidates are evaluated by three independent critics focusing strictly on toxicity, semantic preservation, and fluency. Crucially, the critics are \textit{role-blind}: they evaluate the text without observing which generator produced it, preventing persona bias and evaluator sycophancy.
    \item \textbf{Constraint-First Selection}: Rather than optimizing an unconstrained weighted sum, our selector first enforces hard preservation and fluency feasibility thresholds. Only candidates satisfying these non-negotiable criteria are ranked by utility, giving semantic fidelity veto power over superficial safety.
\end{enumerate}

Through systematic ablation of the iterative loop, we observe that open-ended multi-round debate produces diminishing returns ($+0.0061$ over single-round selection, with confidence intervals crossing zero). We therefore formulate \textbf{AMEND Lite}, a lightweight, bounded control loop. AMEND Lite harnesses the full three-role diversity in the initial round, and invokes at most a single targeted repair round with the minimal editor only if residual toxic spans remain.

We summarize our main contributions as follows:
\begin{itemize}
    \item We propose AMEND, a modular, inference-time framework that decouples text detoxification into role-conditioned proposal generation, isolated role-blind criticism, and constraint-first selection.
    \item We formulate constraint-first selection with hard feasibility gates, preventing low toxicity from masking semantic distortion or disfluency.
    \item We develop AMEND Lite, an efficient, bounded controller that reduces inference overhead by up to $28.4\%$ while boosting detoxification performance over unconstrained multi-round loops.
    \item Across comprehensive, seed-aligned experiments on ParaDetox, we demonstrate that role diversity and constrained selection yield statistically significant improvements ($+0.0508$ over the best single role; $+0.0489$ over legacy multi-round loops), establishing a principled foundation for agentic text revision.
\end{itemize}
\begin{figure*}[t]
  \centering
  \includegraphics[width=\textwidth]{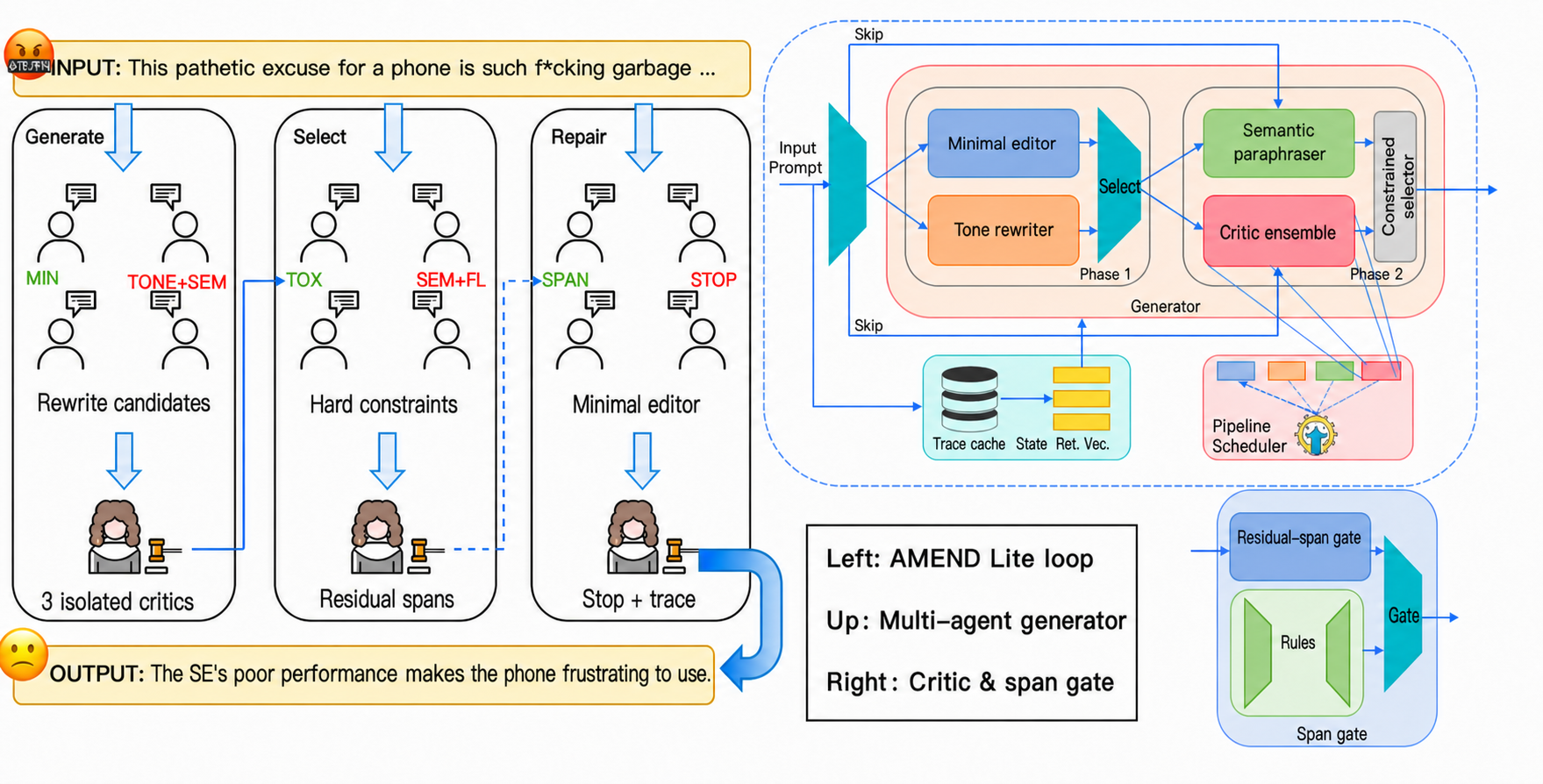}
  \caption{Overview of the AMEND Lite architecture. In Round 1, three role-conditioned generators produce diverse candidates, which are evaluated by isolated, role-blind critics. The constraint-first selector filters candidates by feasibility thresholds before ranking by utility. If residual toxic spans persist, a bounded Round 2 invokes the minimal editor for targeted repair, enforcing monotonic acceptance.}
  \label{fig:pipeline}
\end{figure*}

\section{Related Work}

\subsection{Text Detoxification}

Text detoxification has been investigated through the lenses of controlled text generation and style transfer. \citet{dale-etal-2021-text} established strong baselines using pretrained sequence-to-sequence models, including localized mask-and-replace pipelines and paraphrase generation. \citet{logacheva-etal-2022-paradetox} introduced ParaDetox, a parallel benchmark of over 10,000 toxic English sentences paired with high-quality human neutralizations. This resource was later expanded across Russian, Ukrainian, and Spanish by \citet{dementieva-etal-2024-multiparadetox}.

Recent efforts have explored diverse mechanisms to navigate the safety--preservation trade-off. COUNT applies contrastive unlikelihood training to penalize the generation of toxic identity tokens \citep{pour-etal-2023-count}. MaRCo steers controllable generation using expert and anti-expert language model signals \citep{hallinan-etal-2023-detoxifying}. XDetox leverages token-level attribution explanations to locate toxic spans and rerank rewrites \citep{lee-etal-2024-xdetox}, while DetoxLLM utilizes pseudo-parallel corpora and explanation-guided revision to identify inputs that cannot be detoxified without semantic loss \citep{khondaker-etal-2024-detoxllm}. AMEND departs from these approaches by generating full-sentence proposals across diverse behavioral roles and evaluating them via isolated, dimension-specific critics before constraint-guided selection.

\subsection{Multidimensional Evaluation}

Evaluation in text style transfer fundamentally requires measuring style transfer intensity, content preservation, and naturalness independently \citep{mir-etal-2019-evaluating}. This multidimensionality is especially pronounced in detoxification: naive masking easily achieves perfect toxicity removal at the expense of fluency, whereas unconstrained paraphrasing frequently drifts from the original message. Extensive empirical studies reveal that standard automatic metrics correlate imperfectly with human judgments and display divergent behavior across architectures \citep{logacheva-etal-2022-study}. 

While LLM-based evaluators provide flexible rubric scoring \citep{liu-etal-2023-g-eval,ostheimer-etal-2024-text}, they frequently exhibit position bias, verbosity bias, and self-preference anomalies \citep{zheng-etal-2023-llm-judge}. AMEND directly addresses these challenges by isolating critics to single dimensions, enforcing role blindness, and maintaining explicit decision traces with transparent, reproducible evaluation proxies.

\subsection{Multi-Agent Refinement and Test-Time Compute}

Iterative refinement has emerged as an effective paradigm for improving LLM reasoning and generation without weight updates. Self-Refine demonstrates that a single model can iteratively diagnose and revise its own outputs \citep{madaan-etal-2023-self-refine}. Multi-agent collaboration frameworks introduce persona diversity and interactive deliberation: ReConcile aggregates heterogeneous model outputs through multi-round discussion and confidence-weighted voting \citep{chen-etal-2024-reconcile}, while multi-agent debate protocols leverage competitive interaction to improve factual consistency \citep{du-etal-2024-multiagent-debate}.

However, recent studies caution that multi-agent debate protocols are highly sensitive to prompt design, often suffer from feedback sycophancy, and do not uniformly surpass simple majority voting or ensembling \citep{smit-etal-2024-going-mad}. MAMM-Refine shows that collaborative benefits vary sharply across subtasks, with verification benefiting from isolation \citep{wan-etal-2025-mamm}. Motivated by these findings, AMEND adopts a selection-first philosophy: diverse generator roles propose independent solutions, isolated critics provide objective evidence, and the controller restricts iterative refinement to targeted, bounded repairs.

\section{The AMEND Framework}

\subsection{Task Formulation and the Feasibility-First Principle}

Given an input utterance $x$, the objective of text detoxification is to generate a rewrite $y$ that eliminates toxic expression while strictly preserving the underlying proposition and ensuring fluent surface realization. Let $T(y) \in [0, 1]$ represent toxicity, $S(x, y) \in [0, 1]$ semantic preservation, and $F(y) \in [0, 1]$ fluency.

Standard approaches commonly collapse these dimensions into an unconstrained composite score $U_{\text{linear}} = w_1(1-T) + w_2 S + w_3 F$. However, linear combination permits an evasion trap: a rewrite that completely changes the topic can attain near-zero toxicity and high fluency, achieving an acceptable scalar score despite completely failing the core task.

To eliminate this failure mode, AMEND establishes a \textbf{constraint-first selection} mechanism. We define candidate feasibility $\mathcal{C}(x, y)$ via hard thresholds:
\begin{equation}
\mathcal{C}(x, y) = \mathbb{I}(S(x, y) \ge \tau_S \;\land\; F(y) \ge \tau_F),
\end{equation}
where $\tau_S = 0.72$ and $\tau_F = 0.70$ in our implementation. Among all candidates in the feasible set $\mathcal{Y}_{\text{feas}} = \{y \in \mathcal{Y} \mid \mathcal{C}(x, y) = 1\}$, candidates are ranked by overall utility:
\begin{equation}
\begin{aligned}
U(y) = &\;0.55(1 - T(y)) \\
&+ 0.30 S(x, y) + 0.15 F(y).
\end{aligned}
\end{equation}
This formulation guarantees that semantic fidelity and readability exercise strict veto power before safety optimization occurs.

\subsection{Role-Conditioned Proposal Generation}

Rather than sampling multiple outputs from an unconditioned prompt, the first round of AMEND deploys three specialized generator roles concurrently:
\begin{itemize}
    \item \textbf{Minimal Editor} (\texttt{minimal\_editor}): Replaces or removes toxic spans while strictly preserving original sentence syntax and unaffected lexical items.
    \item \textbf{Tone Rewriter} (\texttt{tone\_rewriter}): Preserves the speaker's stance and communicative action while reframing hostility into firm, assertive, and civil discourse.
    \item \textbf{Semantic Paraphraser} (\texttt{semantic\_paraphraser}): Performs broader structural paraphrasing, explicitly safeguarding entities, factual relations, negation, modality, and requested actions.
\end{itemize}
All roles share the same underlying language model and decoding parameters; strategic diversity is induced solely through role-specific prompt contracts. Before evaluation, candidates are canonicalized: empty strings, duplicate outputs, and previously observed candidates in the execution trace are filtered out.

\subsection{Isolated, Role-Blind Critics}

To ensure objective and unbiased evaluation, each generated candidate is assessed by three independent critic agents:
\begin{itemize}
    \item \textbf{Toxicity Critic}: Estimates residual toxicity $T(y)$ and extracts specific toxic spans remaining in the candidate.
    \item \textbf{Semantic Critic}: Evaluates semantic preservation $S(x, y)$ with respect to source entities, propositional content, negation, modality, and communicative intent.
    \item \textbf{Fluency Critic}: Assesses grammatical correctness, natural phrasing $F(y)$, and absence of unnatural editing artifacts.
\end{itemize}
\textbf{Role Blindness}: Crucially, critics receive only the source text $x$, candidate text $y$, and task language. The critics are completely blind to which generator role produced the candidate. This isolation eliminates persona sycophancy, halo effects, and evaluator bias. Each critic operates under a separate prompt schema and executes concurrently, returning both a scalar score and structured diagnostic rationale.

\subsection{Constraint-First Selection and Diagnostic Fallback}

The selector filters the candidate pool by evaluating feasibility constraints. If $\mathcal{Y}_{\text{feas}} \neq \emptyset$, the candidate maximizing $U(y)$ is selected as the current round state.

If no candidate satisfies the feasibility constraints ($\mathcal{Y}_{\text{feas}} = \emptyset$), the selector does not accept any candidate into the rewrite state. Instead, it identifies the candidate with the minimal aggregate constraint violation:
\begin{equation}
\begin{aligned}
y^*_{\text{diag}} = \arg\min_{y \in \mathcal{Y}} \Big[ &\max(0, \tau_S - S(x, y)) \\
&+ \max(0, \tau_F - F(y)) \Big].
\end{aligned}
\end{equation}
This candidate is logged as a diagnostic repair target for downstream intervention without compromising the integrity of the active state.

\subsection{AMEND Lite: Bounded, Targeted Control}

\begin{figure*}[t]
  \centering
  \includegraphics[width=0.92\textwidth]{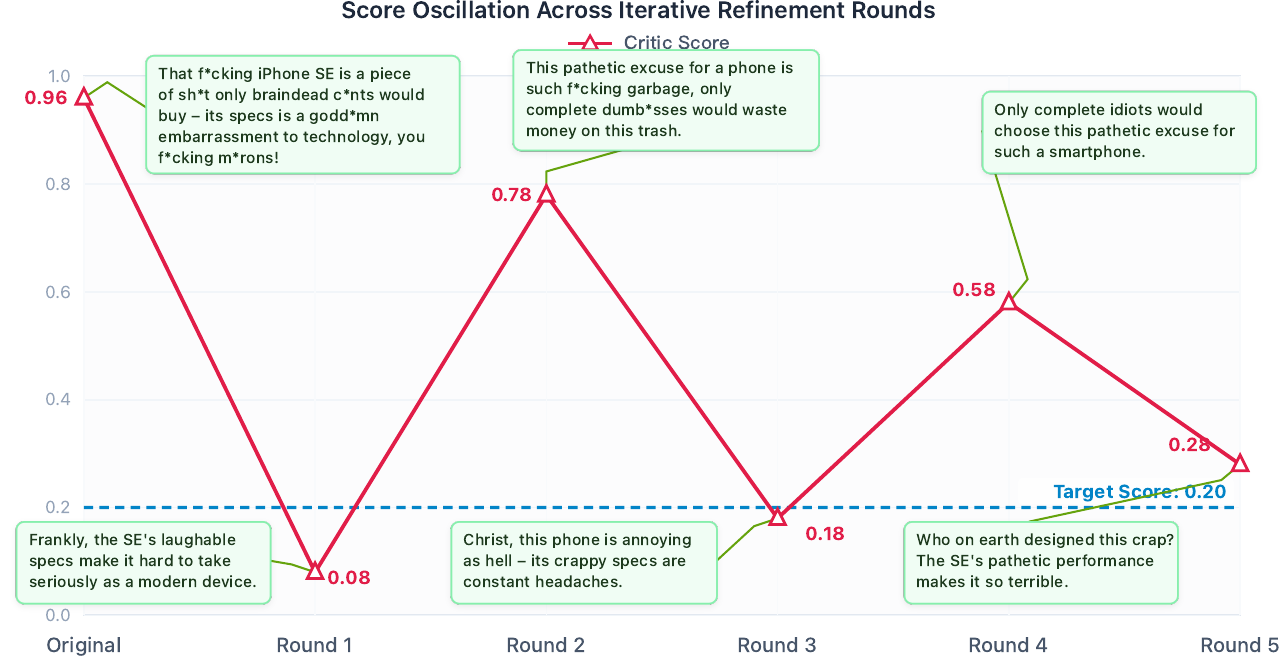}
  \caption{Case study of pathological score oscillation in unconstrained iterative refinement. Rather than converging monotonically toward the safety target ($\tau = 0.20$), the rewrite trajectory repeatedly fluctuates between safe paraphrasing (e.g., Round 1: $T = 0.08$) and severe regression (e.g., Round 2: $T = 0.78$), illustrating why unconstrained multi-round debate produces diminishing returns and motivating AMEND Lite's bounded controller, residual-span gating, and monotonic acceptance rules.}
  \label{fig:oscillation}
\end{figure*}

Figure~\ref{fig:pipeline} illustrates the complete AMEND Lite pipeline. While multi-agent debate can theoretically iterate indefinitely, our empirical analyses demonstrate that open-ended multi-round interaction offers negligible quality improvements while sharply inflating computational cost. In particular, unconstrained iterative refinement frequently suffers from pathological score oscillation (Figure~\ref{fig:oscillation}), where successive rounds alternate between over-correction and re-introducing toxic slurs without monotonic convergence.

AMEND Lite addresses this inefficiency through a bounded two-round control policy:
\begin{enumerate}
    \item \textbf{Initial Search (Round 1)}: The system executes the full three-role proposal and three-critic evaluation pipeline. If the input is already safe, or if the selected candidate achieves the target toxicity threshold and contains zero residual toxic spans, the process terminates immediately.
    \item \textbf{Targeted Repair (Round 2)}: If residual toxic spans remain, the controller triggers a single repair round. To prevent semantic drift, Round 2 invokes exclusively the \texttt{minimal\_editor}, supplied with the specific residual spans identified by the toxicity critic.
    \item \textbf{Monotonic Acceptance}: A repair candidate is accepted if and only if it is feasible and either: (i) reduces toxicity by at least a minimal margin, or (ii) maintains toxicity while strictly decreasing the count of residual toxic spans.
    \item \textbf{Guaranteed Termination}: The process halts upon reaching the safety target, encountering a duplicate state, observing no improvement, or reaching the two-round ceiling.
\end{enumerate}
By restricting multi-role search to Round 1 and focusing Round 2 on surgical repair, AMEND Lite combines exploration diversity with execution efficiency.

\section{Experimental Setup}

\subsection{Dataset and Deterministic Sampling}

We evaluate AMEND on the English ParaDetox benchmark \citep{logacheva-etal-2022-paradetox}, utilizing the standard parallel training split (SHA-256: \texttt{6c305890...9c72ebc}). To ensure reproducible, unconfounded comparisons, we implement a strict deterministic sampling pipeline: source utterances matching an established harmful-term lexicon are deduplicated, ordered deterministically by SHA-256 hash, and assigned contiguous offsets:
\begin{itemize}
    \item \textit{Prompt Development}: Offsets 0--9 (10 inputs) were used solely for prompt drafting and system design.
    \item \textit{Initial Ablation Split}: Offsets 10--109 (100 inputs) evaluate generator role diversity and iterative feedback depth.
    \item \textit{Validation Split}: Offsets 110--139 (30 inputs) were used for tuning control thresholds in AMEND Lite.
    \item \textit{Frozen Confirmation Split}: Offsets 140--239 (100 inputs) serve as a completely untouched, frozen evaluation set for confirming AMEND Lite.
\end{itemize}

\subsection{Model and Decoding Configuration}

All reported experiments utilize a local 4-bit \texttt{qwen3.5-4b-mlx} model executed via an OpenAI-compatible LM Studio runtime on Apple Silicon. Decoding parameters are fixed across all agents and critics: temperature $0.2$, top-$p$ $0.8$, top-$k$ $20$, and a maximum generation budget of 128 tokens. Experiments are repeated across three fixed random seeds: 42, 43, and 44.

\subsection{Evaluation Metrics and Statistical Testing}

To provide objective, reproducible evaluation, we report transparent automatic metrics across all dimensions:
\begin{itemize}
    \item \textbf{Style (Lexicon)}: Binary indicator ($1$ or $0$) denoting the complete absence of matches from the target evaluation toxicity lexicon.
    \item \textbf{Ref-F1}: Unigram token F1 score computed against the ground-truth human neutralization provided by ParaDetox.
    \item \textbf{Source-F1}: Unigram token F1 score computed against the original source input, reflecting lexical and syntactic retention.
    \item \textbf{Surface Fluency}: Heuristic fluency metric that evaluates string validity, regular punctuation, spacing, and penalizes explicit deletion artifacts (e.g., \texttt{[REMOVED]}).
    \item \textbf{Joint Proxy ($J$)}: Following standard practice, overall multi-objective quality is defined at the instance level:
    \begin{equation}
    J_i = \text{Style}_i \times \text{Ref-F1}_i \times \text{Fluency}_i.
    \end{equation}
    The reported Joint score is the sample mean of $J_i$ across instances, rather than a product of aggregate corpus means.
\end{itemize}

\textbf{Statistical Significance}: For model conditions, we report the mean across seeds along with the sample standard deviation ($\pm$). For paired condition comparisons, sample scores are first averaged across the three seeds, followed by 5,000 paired bootstrap iterations over the 100 sample-level differences to construct percentile $95\%$ confidence intervals.

\begin{table*}[t]
\centering
\small
\begin{tabular}{lccccc}
\toprule
\multirow{2}{*}{\textbf{Method / Configuration}} & \textbf{Safety} & \multicolumn{2}{c}{\textbf{Content Preservation}} & \textbf{Fluency} & \textbf{Overall Quality} \\
\cmidrule(lr){2-2} \cmidrule(lr){3-4} \cmidrule(lr){5-5} \cmidrule(lr){6-6}
& Style (Lex.) $\uparrow$ & Ref-F1 $\uparrow$ & Source-F1 $\uparrow$ & Surface $\uparrow$ & Joint Proxy ($J \uparrow$) \\
\midrule
\multicolumn{6}{l}{\textit{Heuristic Baseline}} \\
Keyword Masking & \textbf{1.0000} & 0.7512 & --- & 0.3770 & 0.2817 \\
\addlinespace
\multicolumn{6}{l}{\textit{Single-Role Generators (No Multi-Agent Selection)}} \\
Minimal Editor & 0.5867 & \textbf{0.7778} & \textbf{0.9010} & 0.9607 & $0.4623 \pm 0.0122$ \\
Tone Rewriter & 0.7533 & 0.7102 & 0.7590 & \textbf{0.9850} & $0.5385 \pm 0.0080$ \\
Semantic Paraphraser & 0.6967 & 0.7280 & 0.7912 & 0.9820 & $0.5141 \pm 0.0165$ \\
\addlinespace
\multicolumn{6}{l}{\textit{Constraint-Guided Multi-Agent Framework (Ours)}} \\
AMEND (One-round selection) & 0.8000 & 0.7384 & 0.7983 & 0.9807 & \underline{$0.5894 \pm 0.0245$} \\
AMEND (Full multi-round, $\le 3$ rounds) & \underline{0.8100} & 0.7395 & 0.7979 & 0.9797 & $\mathbf{0.5954 \pm 0.0196}$ \\
\bottomrule
\end{tabular}
\caption{Ablation of generator role diversity and iterative feedback depth on 100 ParaDetox inputs (seeds 42, 43, 44). Style measures lexicon-level detoxification; Ref-F1 and Source-F1 evaluate semantic overlap against human references and source inputs; Surface denotes output fluency; and Joint Proxy ($J$) is the instance-level product $\text{Style} \times \text{Ref-F1} \times \text{Fluency}$. $\pm$ denotes sample standard deviation across seed means. Multi-agent selection over diverse roles yields a $+0.0508$ boost over the strongest single role, capturing $89\%$ of the gains achievable by full iterative debate. Bold denotes best overall; underline denotes second-best.}
\label{tab:ablation}
\end{table*}

\section{Results}

\subsection{Role Diversity and Constrained Selection}

Table~\ref{tab:ablation} presents the initial ablation investigating proposal diversity and selection. The results demonstrate that constraint-guided selection across diverse roles provides substantial, statistically robust advantages over any individual role.

Specifically, single-round selection across the three generator roles achieves a Joint proxy score of $0.5894 \pm 0.0245$, substantially outperforming the strongest single-role condition (\texttt{tone\_rewriter} at $0.5385 \pm 0.0080$). The paired improvement is $+0.0508$, with a $95\%$ bootstrap confidence interval of $[+0.0224, +0.0812]$, which excludes zero by a wide margin. Furthermore, one-round AMEND improves upon the minimal editor by $+0.1270$ ($95\%$ CI $[+0.0778, +0.1789]$) and the semantic paraphraser by $+0.0752$ ($95\%$ CI $[+0.0341, +0.1186]$).

\textbf{Proposal Distribution}: An analysis of candidate selections confirms that all three roles contribute meaningfully to the final output. Across 276 non-safe first-round decisions, the \texttt{minimal\_editor} was chosen 157 times ($56.9\%$), the \texttt{tone\_rewriter} 99 times ($35.9\%$), and the \texttt{semantic\_paraphraser} 20 times ($7.2\%$). Each non-safe input yielded an average of 2.36 unique candidates. This distribution highlights that conservative localized editing provides the primary foundation, while tone reframing and semantic paraphrasing provide vital alternatives when simple lexical substitution fails feasibility constraints.

\begin{table*}[t]
\centering
\small
\begin{tabular}{lccccc}
\toprule
\multirow{2}{*}{\textbf{Method / Configuration}} & \textbf{Safety} & \multicolumn{2}{c}{\textbf{Content Preservation}} & \textbf{Fluency} & \textbf{Overall Quality} \\
\cmidrule(lr){2-2} \cmidrule(lr){3-4} \cmidrule(lr){5-5} \cmidrule(lr){6-6}
& Style (Lex.) $\uparrow$ & Ref-F1 $\uparrow$ & Source-F1 $\uparrow$ & Surface $\uparrow$ & Joint Proxy ($J \uparrow$) \\
\midrule
Keyword Masking & \textbf{1.0000} & 0.7449 & --- & 0.3710 & 0.2761 \\
Legacy AMEND (Up to 3 rounds) & 0.8167 & 0.7497 & 0.7941 & 0.9860 & $0.6108 \pm 0.0156$ \\
\midrule
\textbf{AMEND Lite} (Targeted 2-round repair) & \underline{0.8767} & \textbf{0.7582} & \textbf{0.8067} & \textbf{0.9870} & $\mathbf{0.6597 \pm 0.0062}$ \\
\bottomrule
\end{tabular}
\caption{Frozen confirmation results on a disjoint 100-example test split of ParaDetox (offsets 140--239, seeds 42, 43, 44). AMEND Lite achieves simultaneous improvements in safety ($+6.0\%$ in Style) and content preservation ($+0.85\%$ Ref-F1, $+1.26\%$ Source-F1) without compromising fluency, resulting in a $+0.0489$ increase in Joint score over legacy AMEND ($95\%$ bootstrap CI $[+0.0212, +0.0788]$). Bold denotes best model performance; underline denotes highest safety among generative methods.}
\label{tab:confirmation}
\end{table*}

\subsection{Frozen Confirmation of AMEND Lite}

Table~\ref{tab:confirmation} reports the performance of AMEND Lite on the untouched frozen confirmation split (offsets 140--239). AMEND Lite raises the Joint proxy score to $0.6597 \pm 0.0062$, compared to $0.6108 \pm 0.0156$ for legacy AMEND and $0.2761$ for keyword masking.

Importantly, AMEND Lite avoids trading semantic preservation for detoxification: Style increases from $0.8167$ to $0.8767$ ($+6.0\%$), while Ref-F1 increases from $0.7497$ to $0.7582$ and Source-F1 increases from $0.7941$ to $0.8067$. Surface fluency remains uniformly high ($0.9870$ vs. $0.9860$). 

In paired bootstrap analysis, the paired difference between AMEND Lite and legacy AMEND is $+0.0489$, with a $95\%$ confidence interval of $[+0.0212, +0.0788]$. Lite improves over keyword masking by $+0.3836$ (CI $[+0.3286, +0.4360]$). Across the three independent seeds, AMEND Lite scores remain remarkably stable ($0.6661$, $0.6536$, and $0.6596$), demonstrating that the bounded controller provides reliable, reproducible performance gains.

\subsection{Feedback Depth and Iterative Saturation}

Comparing the legacy multi-round controller with single-round multi-agent selection in Table~\ref{tab:ablation} reveals an essential finding: extending the full multi-agent loop for up to three rounds produces only a marginal gain of $+0.0061$, with a $95\%$ bootstrap confidence interval of $[-0.0028, +0.0163]$ that crosses zero.

Out of 300 seed-level execution traces under the legacy controller, 243 terminated early upon reaching the safety target, 24 bypassed generation via the safe-input path, 11 stopped due to lack of progress, and 22 were halted by cycle detection; exactly zero runs terminated unresolved at the maximum round limit. These dynamics prove that open-ended multi-agent debate quickly saturates, strongly motivating AMEND Lite's bounded design of dedicating compute to first-round proposal diversity and restricting subsequent iterations to surgical repair.

\begin{table}[t]
\centering
\small
\setlength{\tabcolsep}{3pt}
\begin{tabular}{lrrr}
\toprule
\textbf{Computational Metric} & \textbf{Legacy} & \textbf{Lite} & \textbf{$\Delta$ (\%)} \\
\midrule
\multicolumn{4}{l}{\textit{Inference Calls}} \\
\quad Physical requests / item & 13.19 & 9.45 & $-28.4\%$ \\
\quad Standalone est. requests & 13.19 & 12.45 & $-5.6\%$ \\
\addlinespace
\multicolumn{4}{l}{\textit{Token Consumption}} \\
\quad Prompt tokens / item & 3466.7 & 2481.4 & $-28.4\%$ \\
\quad Completion tokens / item & 654.2 & 439.7 & $-32.8\%$ \\
\addlinespace
\multicolumn{4}{l}{\textit{Latency \& Search Dynamics}} \\
\quad Latency / item (s) & 54.6 & 38.0 & $-30.4\%$ \\
\quad Execution rounds / item & 1.087 & 1.007 & $-7.4\%$ \\
\quad Candidates / item & 2.31 & 2.10 & $-9.2\%$ \\
\bottomrule
\end{tabular}
\caption{Computational profiling and execution audit on the confirmation subset. AMEND Lite substantially curtails token overhead and latency by eliminating redundant search rounds. In paired evaluation, Lite reuses input critiques (yielding a $28.4\%$ request reduction); restoring these critiques in standalone deployment still yields an overall efficiency gain.}
\label{tab:cost}
\end{table}

\subsection{Computational Profiling and Cost Audit}

Table~\ref{tab:cost} presents an audit of computational efficiency across 600 execution traces and 6,793 model invocations. In the physical paired run, AMEND Lite achieves a $28.4\%$ reduction in model requests per item ($9.45$ vs. $13.19$), a $28.4\%$ decrease in prompt tokens, and a $32.8\%$ reduction in completion tokens, translating to a $30.4\%$ drop in wall-clock latency ($38.0$s vs. $54.6$s per example).

In standalone deployment without shared input caching, AMEND Lite requires an estimated $12.45$ requests per item, maintaining a $5.6\%$ request advantage over legacy AMEND while eliminating the risk of degenerative looping.

\section{Analysis}

\subsection{Why Proposal Diversity Outperforms Iteration}

Our empirical findings clarify the mechanics of test-time collaboration in text generation. The significant performance leap occurs when transitioning from a single prompt persona to three diverse roles ($+0.0508$), whereas iterating the full multi-agent dialogue yields minimal benefit ($+0.0061$). 

This asymmetry stems from the nature of the detoxification search space. A single generator role is inherently constrained: the minimal editor cannot easily reframe toxic sarcasm, while the tone rewriter occasionally over-corrects. Generating candidates across orthogonal perspectives exposes a rich frontier of viable alternatives. The constraint-first selector then identifies the optimal candidate immediately. Subsequent multi-agent dialogue often generates minor lexical perturbations that rarely overcome the initial selections, while risking hallucination and semantic drift.

\subsection{Mechanisms of AMEND Lite}

AMEND Lite succeeds because it aligns test-time compute with empirical return. By reserving the full ensemble search for Round 1, it captures maximum candidate diversity upfront. When residual toxicity persists, rather than reigniting an open-ended debate, it executes a single targeted repair with the minimal editor conditioned on critic-extracted spans. Combined with duplicate filtering, monotonic acceptance gates, and cycle detection, AMEND Lite achieves superior detoxification quality while drastically reducing inference cost.

\section{Conclusion}

In this work, we presented AMEND, an inference-time framework that structures text detoxification into role-conditioned proposal generation, isolated multidimensional criticism, and constraint-first selection. AMEND replaces monolithic generation with structured collaboration: complementary generator roles explore the edit space, role-blind critics evaluate candidates without bias, and hard feasibility constraints prevent safety from compromising semantic preservation. 

Furthermore, we introduced AMEND Lite, a bounded controller that concentrates multi-role search in the first round and restricts subsequent computation to pinpoint repairs. Across systematic experiments on ParaDetox, AMEND Lite achieves significant gains in safety and content fidelity while cutting inference requests by up to $28.4\%$. Our findings suggest that in agentic generation, structured initial diversity coupled with rigorous feasibility filtering is far more effective than unconstrained iterative debate, offering a clear path toward efficient, controllable text revision.

\section*{Limitations}

While AMEND demonstrates clear empirical advantages in controllable text detoxification, several dimensions define the scope of the present work:
\begin{itemize}
    \item \textbf{Model Scale and Diversity}: Experiments were conducted using a quantized local \texttt{qwen3.5-4b-mlx} model. While this demonstrates that AMEND can unlock strong performance even with compact, consumer-accessible models, evaluating our framework across larger model families (e.g., Qwen-72B, LLaMA-3) and proprietary APIs represents a natural next step.
    \item \textbf{Metric Scope}: Evaluation relies on transparent, reproducible automatic proxies (lexicon-based style, unigram token F1, and heuristic surface fluency). While these provide standardized, reproducible benchmarks, they do not fully capture implicit toxicity or nuanced human stylistic preferences. Incorporating trained neural evaluators and blinded human judgments will provide valuable complementary validation.
    \item \textbf{Linguistic Coverage}: Our experiments focused on English ParaDetox. Extending AMEND to multilingual settings (such as MultiParaDetox) will further test how role-conditioned personas generalize across diverse linguistic and cultural norms.
\end{itemize}

\section*{Ethical Considerations}

Text detoxification tools can facilitate healthier online discourse and assist moderation systems, but automated rewriting must be applied with care. Preserving authorial stance, identity references, and factual assertions is critical to avoid covert censorship or erasing legitimate dissent. AMEND's transparent constraint-first selection and auditable decision traces enhance human oversight, helping ensure that detoxification does not distort meaning or suppress authentic communication.

\section*{Code and Data Availability}

The complete source code, test suites, deterministic sampling routines, and configuration files will be released soon (coming soon) at \url{https://github.com/Clearwavelet/AMEND}.

\section*{Author Contributions}

\textbf{Tianyu Wang} conceived the project, designed and implemented the AMEND framework, conducted the experiments, analyzed the empirical data, and wrote the manuscript.

\section*{Conflict of Interest}

The author declares no conflict of interest.

\section*{Funding}

This research received no external funding.

\section*{Acknowledgments}

The author acknowledges OpenAI Codex for assistance with manuscript drafting, LaTeX typesetting, and literature organization. The author verified all citations and numerical results against experimental traces and takes full responsibility for the contents of this paper.

\bibliography{references}

\appendix

\section{Implementation and Verification Details}

AMEND represents each execution trace as a structured, strongly-typed record logging input and output critiques, per-round candidate pools, feasibility determinations, utility scores, selected texts, execution status, and iteration counts. Offline unit tests verify safe-input bypass, multilingual control flow, residual-span gating, duplicate filtering, targeted second-round repair, and metric aggregation. At the time of report compilation, all 29 tests passed under Python 3.9.6 and Python 3.13.12, ensuring framework stability and reproducibility.

\section{Paired Statistical Comparisons}

Table~\ref{tab:paired_comparisons} provides the complete paired bootstrap analysis across all evaluated conditions on the 100-example subsets. All comparisons between multi-agent selection and single-role baselines exclude zero, confirming the statistical significance of role diversity. In contrast, the comparison between full multi-round iteration and one-round selection crosses zero, empirically confirming the saturation of open-ended debate.

\begin{table}[b]
\centering
\small
\setlength{\tabcolsep}{3pt}
\begin{tabular}{lrc}
\toprule
\textbf{Comparison} & \textbf{Diff.} & \textbf{95\% Bootstrap CI} \\
\midrule
\multicolumn{3}{l}{\textit{vs. Heuristic Baseline}} \\
One-round $-$ Masking & $+0.3077$ & $[+.2488, +.3656]^*$ \\
Full $-$ Masking & $+0.3138$ & $[+.2552, +.3711]^*$ \\
Lite $-$ Masking & $+0.3836$ & $[+.3286, +.4360]^*$ \\
\addlinespace
\multicolumn{3}{l}{\textit{Role Diversity vs. Single Roles (Ablation)}} \\
One-round $-$ Minimal & $+0.1270$ & $[+.0778, +.1789]^*$ \\
One-round $-$ Tone & $+0.0508$ & $[+.0224, +.0812]^*$ \\
One-round $-$ Semantic & $+0.0752$ & $[+.0341, +.1186]^*$ \\
\addlinespace
\multicolumn{3}{l}{\textit{Feedback Loop Saturation}} \\
Full $-$ One-round & $+0.0061$ & $[-.0028, +.0163]$\phantom{$^*$} \\
\addlinespace
\multicolumn{3}{l}{\textit{Controller Optimization (Confirmation)}} \\
Lite $-$ Legacy & $+0.0489$ & $[+.0212, +.0788]^*$ \\
\bottomrule
\end{tabular}
\caption{Sample-aligned paired bootstrap comparisons across 5,000 resamples. Differences are calculated on instance-level Joint proxy means averaged across seeds (42, 43, 44). Asterisk ($*$) indicates statistical significance ($95\%$ bootstrap confidence interval strictly excludes zero).}
\label{tab:paired_comparisons}
\end{table}

\end{document}